\documentclass{article}
\usepackage{spconf,amsmath,graphicx, cite, amssymb, amsmath, booktabs, bm}

\title{Improving Few-shot Learning with Weakly-supervised Object Localization}
\name{Inyong Koo, \qquad Minki Jeong, \qquad Changick Kim}
\address{Korea Advanced Institute of Science and Technology (KAIST)\\
School of Electrical Engineering\\
Daejeon, Republic of Korea
}
\begin{document}
%
\maketitle
\begin{abstract}
Few-shot learning often involves metric learning-based classifiers, which predict the image label by comparing the distance between the extracted feature vector and class representations. However, applying global pooling in the back-end of the feature extractor may not produce an embedding that correctly focuses on the class object. In this work, we propose a novel framework that generates class representations by extracting features from class-relevant regions of the images. Given only a few exemplary images with image-level labels, our framework first localizes the class objects by spatially decomposing the similarity between the images and their class prototypes. Then, enhanced class representations are achieved from the localization results. We also propose a loss function to enhance distinctions of the refined features. Our method outperforms the baseline few-shot model in \textit{mini}ImageNet and \textit{tiered}ImageNet benchmarks. 
\end{abstract}
\section{Introduction}
\label{sec:intro}

Deep learning models based on Convolutional Neural Networks (CNNs) have shown remarkable performance in the image classification task \cite{simonyan2015vggnet, he2016resnet, hu2018senet}. These models are usually trained in a supervised manner, relying on a large-scale dataset that provides sufficient labeled image samples. However, in practice, collecting and labeling thousands of target class images would be expensive, sometimes impossible. Few-shot learning addresses the classification problems in the limited conditions, where the information of each class is given with only a few exemplary (or \textit{support}) images.

In a few-shot setting, the scarcity of sample data makes conventional classifiers with a fully-connected layer vulnerable to overfitting. Hence metric learning-based classifiers are preferred instead, where a \textit{query} image is mapped to feature space and then classified by the distances to the class representations acquired from support images \cite{koch2015siamese, vinyals2016matching, snell2017prototypical, oreshkin2018tadam, zhang2019variational, simon2020adaptive, ye2020fewshot}. The Siamese Network architecture \cite{koch2015siamese} is the first to validate the importance of feature embedding in one-shot learning. Vinyals \textit{et al.} \cite{vinyals2016matching} propose a training scheme repeating ``episodes'' of evaluation scenarios as a form of meta-learning. Prototypical Networks \cite{snell2017prototypical} introduce a simple classifier using the Euclidean distance to the class \textit{prototypes}, defined as the centroids of each class's support feature vectors. More recent works explore different class representations and distance metrics. Oreshkin \textit{et al.} \cite{oreshkin2018tadam} suggest a task-dependent adaptive metric (TADAM). Zhao \textit{et al.} \cite{zhang2019variational} propose a variational Bayesian framework that represents each class as a distribution. It uses a probabilistic-based metric, which belongs to weighted Euclidean distances. Deep subspace networks (DSN) \cite{simon2020adaptive} utilize subspace representations and classifies a query based on the shortest distance from the query to its projections onto subspaces.

\begin{figure}[t]
\begin{minipage}[b]{.48\linewidth}
  \centering
  \centerline{\includegraphics[width=4cm]{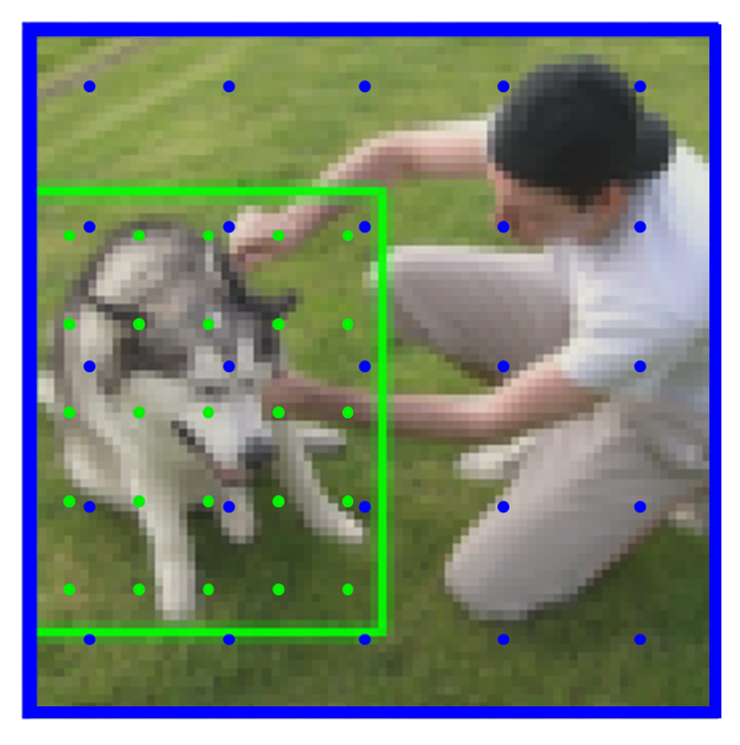}}
  \centerline{(a)\label{fig:1a}}\medskip
\end{minipage}
\hfill
\begin{minipage}[b]{0.48\linewidth}
  \centering
  \centerline{\includegraphics[width=4cm]{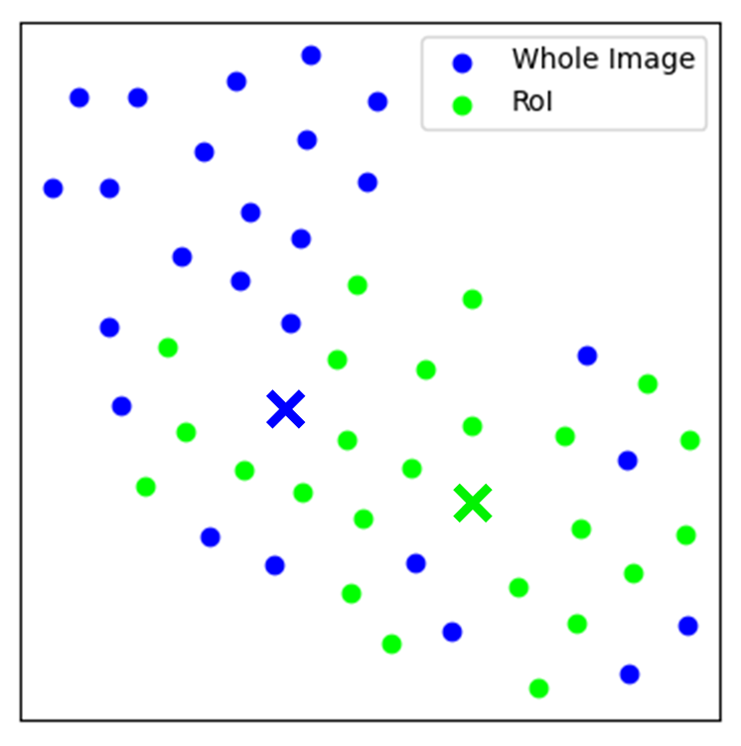}}
  \centerline{(b)\label{fig:1b}}\medskip
\end{minipage}
\vspace{-0.4cm}
\caption{A support image may contain class-irrelevant instances. Our method localizes the class object and acquires the region of interest (RoI) feature. (a) An image of a man and a dog labeled `dog.' (b) The t-SNE plot of features of the whole image and the RoI, colored in blue and green respectively. (`$\times$' denotes the average feature.) }

\label{fig:1}
\end{figure}

\begin{figure*}[htb]
  \centering
  \includegraphics[width=0.95\linewidth]{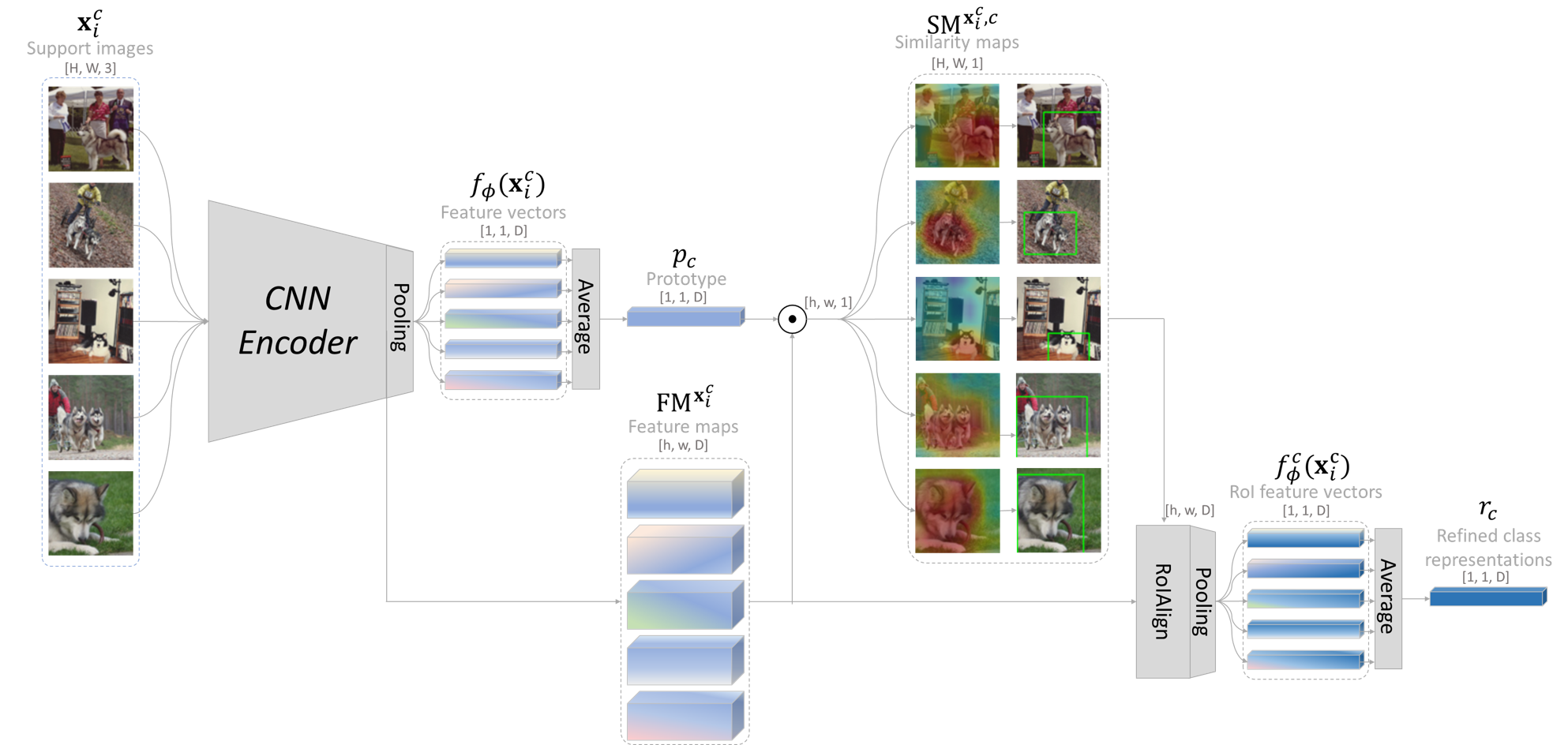}
\caption{Proposed framework. Using the class prototype, we localize the class object and extract a class-relevant feature from each support image to compute refined class representation.  $\odot$ denotes the similarity mapping operation in Eq. \ref{eq:4}.}
\label{fig:2}
\end{figure*}
In metric learning, embedding quality plays a crucial role \cite{tian2020rethink}. An image representation vector is conventionally achieved with a CNN encoder followed by a global average pooling (GAP) layer. However, GAP does not take into account the spatial information of where the feature is originated. As a result, the embedding may contain features of class-irrelevant instances or background. For instance, Fig.~\ref{fig:1a}(a) shows an exemplary image labeled as `dog,' but the extracted feature represents the overall scene of a man with a dog. Figure~\ref{fig:1b}(b) shows that this causes an undesirable bias from the feature that focuses on the foreground object. A class representation obtained from such impure support features may not express the class specifically. To tackle this problem, we propose a framework that localizes the class object and extracts only relevant features at the region of interest (RoI). Our method does not require ground-truth bounding box annotations and localizes the class objects with weak supervision.

A naive approach to address a weakly supervised object localization (WSOL) problem is to use class activation mapping (CAM) \cite{zhou2016learning}. It uses the linear (i.e., fully connected) classifier's weights to visualize activations focusing on the most discriminative parts in the images, consequently highlighting the class objects. Unfortunately, CAM cannot be directly employed for WSOL in few-shot realm due to the overfitting issue of the linear classifier. As a countermeasure, we utilize similarity mapping \cite{stylianou2019visualizing}, which is originally proposed to visualize the similar regions between two images.

Our framework spatially decomposes the similarity between the image features and the class prototypes to localize foreground objects. More class-relevant features are obtained from the predicted bounding boxes via RoIAlign-based feature propagation, resulting in refined class representations. We also introduce a novel loss function that regards our class representations to improve the model's discriminative power. Our method outperforms prior metric learning-based classifiers that do not consider class object locality in \textit{mini}ImageNet \cite{vinyals2016matching} and \textit{tiered}ImageNet \cite{ren18fewshotssl} datasets. The proposed framework also produces good localization results.

\section{Method}
\label{sec:method}

\subsection{Preliminaries}
\label{ssec:protonet}
An $N$-\textit{way} $K$-\textit{shot} problem is a few-shot classification task, where a small support set $S=\{(\textrm{\textbf{x}}_i, y_i)\}_{i=1}^{K}$ is given each for $N$ classes. Here, each $\textrm{\textbf{x}}_i \in \mathbb{R}^{H \times W \times 3}$ is an RGB image with height $H$ and width $W$, and $y_i \in \{1, \cdots, N\}$ is the corresponding label. $S_c$ denotes the support set labeled with class $c$, while $\textrm{\textbf{x}}_i^c$ denotes its support images. 

Prototypical network (ProtoNet) \cite{snell2017prototypical} computes a $D$-dimensional vector $p_c$ as the prototype for each class using an embedding function $f_\phi : \mathbb{R}^{H \times W \times 3} \rightarrow \mathbb{R}^{D}$ with learnable parameters $\phi$. A prototype is the mean vector of the embeddings belonging to its class:
\begin{equation} \label{eq:1}
    p_c = \frac{1}{|S_c|}\sum_{\textrm{\textbf{x}}_i^c \in S_c} f_\phi (\textrm{\textbf{x}}_i^c).
\end{equation}
ProtoNet classifies a query image $\textrm{\textbf{x}}^q$ as the class of the nearest neighboring prototype in the Euclidean distance metric:
\begin{equation} \label{eq:2}
    \hat{y}^q = \min_c \|f_\phi (\textrm{\textbf{x}}^q) - p_c \|.
\end{equation}
\subsection{Proposed framework}
\label{ssec:framework}

Figure \ref{fig:2} shows an overview of our framework. We adopt the ProtoNet architecture to produce `prototype' class representations. During the embedding process, a feature map before back-end spatial pooling layers $\textrm{FM}^{\textrm{\textbf{x}}}  \in \mathbb{R}^{h \times w \times D}$ is retrieved for each image $\textrm{\textbf{x}}$, where $h$ and $w$ denote the channels' height and width.

Class objects are commonly visible among the support images. Thus, each support image would have regions highly correlated with the class representation. After the prototypes are derived, we can compute the similarity map $\textrm{SM}^{\textrm{\textbf{x}}, c} \in \mathbb{R}^{h \times w \times 1}$ to highlight the class-relevant regions of each image by projecting its class prototype to every feature map coordinate:
\begin{align} \label{eq:4}
    \textrm{SM}^{\textrm{\textbf{x}}, c}_{(i,j)} = \frac{p_c}{\|p_c\|} \cdot \textrm{FM}^{\textrm{\textbf{x}}}_{(i,j)} && \textrm{for } (i,j) = (1,1), \cdots, (h,w).
\end{align}
$\textrm{SM}^{\textrm{\textbf{x}}, c}$ is a scaled spatial decomposition of the cosine similarity between the image embedding and the prototype \cite{stylianou2019visualizing}. Next, we upsample similarity maps to the original image size using bilinear interpolation and segment the regions of which the value is above a relative threshold $\tau$ of the max value of each similarity map. Then we take a bounding box that covers the largest connected component.

For every support image, we extract the RoI feature vector $f_\phi^c (\textrm{\textbf{x}}_i^c)$ from the bounding box of its corresponding class via RoIAlign \cite{he2017mask} and pooling operations. We define a novel class representation $r_c$ as the mean vector of the RoI features belonging to its class. Our framework follows the classifying strategy in Eq. \ref{eq:2}, replacing $p_c$ with our new representation $r_c$.

\subsection{Loss function}
\label{ssec:lossfunction}
In ProtoNet, the logits are defined with the squared Euclidean distance to the prototypes and a temperature parameter $T$. The embedding function is optimized with a cross-entropy loss: 

\begin{equation}
    L_{base} (y^q=c | \textrm{\textbf{x}}^q) = -\log \frac{\exp{(-\|f_\phi (\textrm{\textbf{x}}^q) - p_c\|^2 / T)}}{\sum_{k=1}^{N} \exp{(-\|f_\phi (\textrm{\textbf{x}}^q) - p_k\|^2 / T)}} .
\end{equation}

Similarly, we propose a loss term $L_{roi}$ that considers our novel class representations instead of the prototypes:
\begin{equation}
    L_{roi} (y^q=c | \textrm{\textbf{x}}^q) = -\log \frac{\exp{(-\|f_\phi (\textrm{\textbf{x}}^q) - r_c\|^2 / T)}}{\sum_{k=1}^{N} \exp{(-\|f_\phi (\textrm{\textbf{x}}^q) - r_k\|^2 / T)}}.
\end{equation}
$L_{roi}$ has an explicit objective of distinguishing RoI features and allows the model to have more discriminative ability.
Our final loss function is as follows:

\begin{equation}
    L_{ours} = L_{base} + \lambda_{roi} L_{roi},
\end{equation}
where $\lambda_{roi}$ is a balancing parameter.

\section{Experiments}
\label{sec:typestyle}

\subsection{Setups}
\label{ssec:setups}

We evaluated our framework on popular subsets of ImageNet \cite{deng2009imagenet} in few-shot learning studies; \textit{mini}ImageNet \cite{vinyals2016matching} and \textit{tiered}ImageNet \cite{ren18fewshotssl} datasets. \textit{Mini}ImageNet includes 600 images for all 100 classes (i.e. total 60,000 images). We followed the conventional split by \cite{Sachin2017} and used 64 classes for training, 16 classes for validation, and 20 classes for testing. \textit{Tiered}ImageNet is a large-scale dataset which considers the high-level category separation between training and testing classes. \textit{Tiered}ImageNet consists of 351, 97, and 160 classes for training, validation, and testing, respectively.

For both datasets, we evaluated our approach on 5-way 1-shot and 5-way 5-shot scenarios. We report the average classification accuracy and the 95\% confidence interval after performing evaluation on 10,000 sampled tasks. Each task tests 15 query images per class (i.e. 75 queries in total).

\subsection{Implementation details}
\label{ssec:subhead}

Our framework is built on the premise that the prototypes already have reliable representative power, so we trained the ProtoNet to its extent. We selected the ResNet-12 encoder used in \cite{lee2019meta} as our embedding function and initialized it following the pre-training strategy suggested in \cite{qiao2018few, rusu2019meta-learning}. We appended a single layer classifier and trained it for the whole train-split classes using cross-entropy loss to initialize the embedding network's weights. Then we further optimized the feature extractor for 200 epochs with $L_{base}$ using the episodic training scheme \cite{vinyals2016matching}. To be specific, we repeated random batches of few-shot classification tasks (the same scenarios used in evaluation) and trained parameters using SGD optimizer with the learning rate of 0.0001. The temperature parameter $T$ was set to 64, except the case in 5-shot scenario on \textit{tiered}ImageNet where we used $T$= 32. 

Our framework proposes the class object bounding box given $\tau$ = 0.5. We further optimized the encoder's parameters with $L_{ours}$, setting $\lambda_{roi}$= 1.0, 0.5 for 1-shot and 5-shot scenarios, respectively. We used the $\tau$ value of 0.7 instead of 0.5 while training 5-shot scenarios. Additional training takes 100 epochs, with the other hyperparameters unchanged.

\section{Results}
\label{sec:majhead}

\begin{table*}[h]
\centering
\caption{Few-shot classification results and 95\% confidence intervals on \textit{mini}ImageNet and \textit{tiered}ImageNet.}
\label{table:1}
\begin{tabular}[t]{lcccc}
\toprule
& \multicolumn{2}{c}{\textbf{\textit{mini}ImageNet}} & \multicolumn{2}{c}{\textbf{\textit{tiered}ImageNet}}\\
\cmidrule(lr){2-3} \cmidrule(lr){4-5}
&\textbf{5-way 1-shot (\%)}&\textbf{5-way 5-shot (\%)} & \textbf{5-way 1-shot (\%)} & \textbf{5-way 5-shot (\%)} \\
\midrule
TADAM* \cite{oreshkin2018tadam} & 58.50 $\pm$ 0.30 & 76.70 $\pm$ 0.30 & -- & -- \\
Variational FSL* \cite{zhang2019variational} & 61.23 $\pm$ 0.26 & 77.69 $\pm$ 0.17 & -- & -- \\
DSN* \cite{simon2020adaptive} & 62.64 $\pm$ 0.66 & 78.63 $\pm$ 0.45 & 66.22 $\pm$ 0.75 & 82.79 $\pm$ 0.48 \\
CAN* \cite{hou2019cross} & 63.85 $\pm$ 0.48 & 79.44 $\pm$ 0.34 & \textbf{69.89} $\bm{\pm}$ \textbf{0.51} & 84.23 $\pm$ 0.37 \\
\midrule
\multicolumn{3}{l}{\textbf{Optimized with $\bm{L_{base}}$}} \\
ProtoNet \cite{snell2017prototypical} & 63.02 $\pm$ 0.21 & 80.48 $\pm$ 0.14 & 67.76 $\pm$ 0.23 & 84.77 $\pm$ 0.16 \\
Ours   & 63.37 $\pm$ 0.21 & 80.68 $\pm$ 0.14 & 68.06 $\pm$ 0.23 & 84.91 $\pm$ 0.16 \\
\midrule

\multicolumn{3}{l}{\textbf{Optimized with $\bm{L_{ours}}$}} \\
ProtoNet  & 64.35 $\pm$ 0.21 & 81.15 $\pm$ 0.14 & 68.42 $\pm$ 0.23 & 84.86 $\pm$ 0.16 \\
Ours  & \textbf{64.86} $\bm{\pm}$ \textbf{0.21} & \textbf{81.25} $\bm{\pm}$ \textbf{0.14} & 68.87 $\pm$ 0.23 & \textbf{85.07} $\bm{\pm}$ \textbf{0.16} \\
\bottomrule
\multicolumn{5}{r}{\small{* Results reported by the original work.}}
\end{tabular}
\vspace{-0.2cm}
\end{table*}%

\subsection{Classification results comparison}
\label{ssec:QuanRes}
We compared our method with metric learning-based models that use different class representations, all using the ResNet-12 backbone structure. ProtoNet \cite{snell2017prototypical}, our baseline work, shares the same weights with our framework. We also compared with cross attention network (CAN) \cite{hou2019cross}, which shares a similar motivation to ours. CAN extracts a feature that attends to the target object by inspecting the spatial correlation between the query and support images.

Table \ref{table:1} shows the classification performance comparison of our method on the \textit{mini}ImageNet and \textit{tiered}ImageNet. Our framework produces more discriminative class representations than the prototypes. Further optimization using the proposed loss function also enhanced the classification performances for both ProtoNet and our framework. This implies that training with the consideration of distinguishing RoI features improves the overall embedding quality. Our method outperformed prior approaches, except the case in the 5-way 1-shot task on \textit{tiered}ImageNet. CAN exploits the query image to identify the class object, and thus have strength in 1-shot scenarios. Our method produces a query-independent class representation that explicitly localizes and focuses on the target object.

\vspace{-0.2cm}
\subsection{Localization results analysis}
\label{ssec:QualRes}
\vspace{-0.2cm}
\begin{figure}[hb]
  \centering
  \centerline{\includegraphics[width=0.9\linewidth]{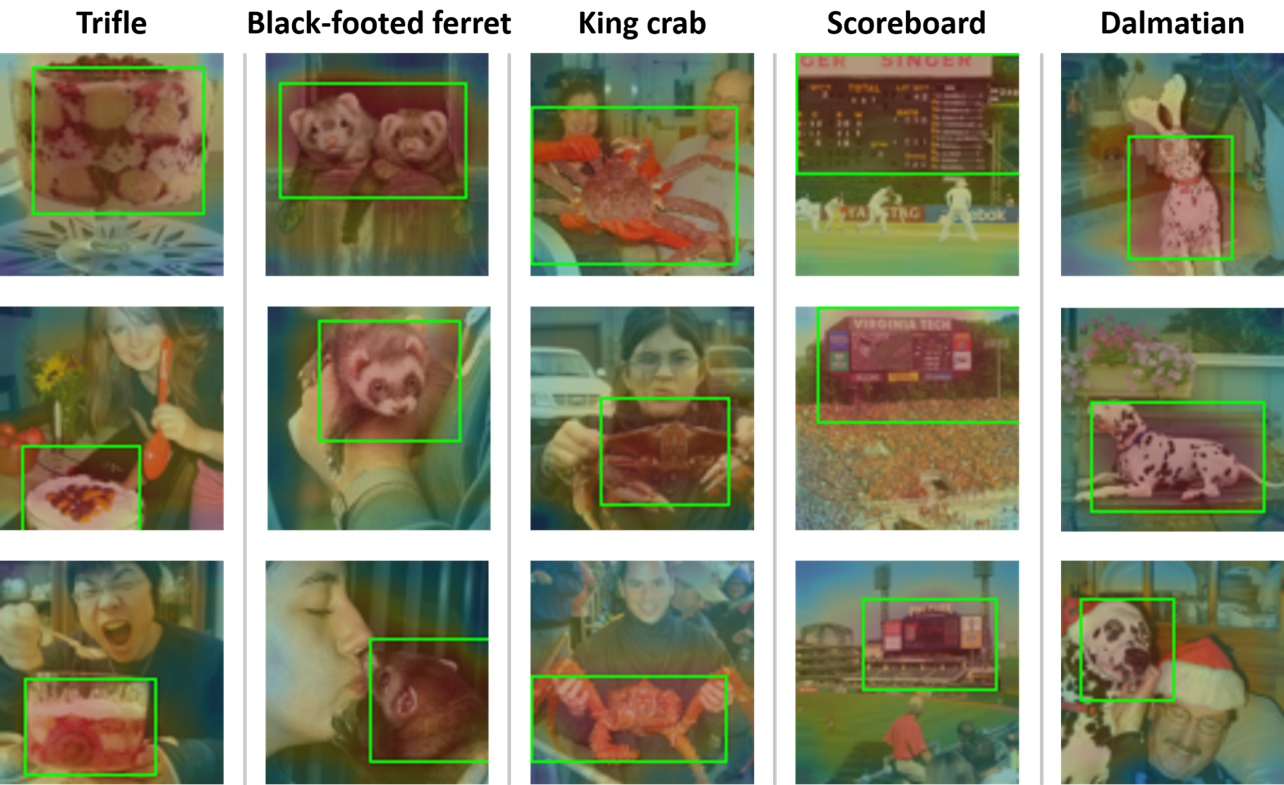}}
\vspace{-0.2cm}
\caption{Localization the class objects from query images in a \textit{mini}ImageNet 5-way 5-shot scenario.}
\label{fig:3}
\end{figure}

Apart from the classification performance, we were concerned about our framework's localizing ability. Our framework produces WSOL predictions as intermediate outcomes. Once we predict a query label, we can also locate the most probable region of the class object using the similarity map between the query feature map and the representation vector of the predicted class. Since neither \textit{mini}ImageNet nor \textit{tiered}ImageNet provide ground-truth bounding box annotations, we were not able to assess the localization quality numerically. However, qualitative results in Fig. \ref{fig:3} show that our framework  can successfully localize the class objects even when there are multiple salient objects.

\begin{figure}[htb]
  \centering
  \centerline{\includegraphics[width=0.9\linewidth]{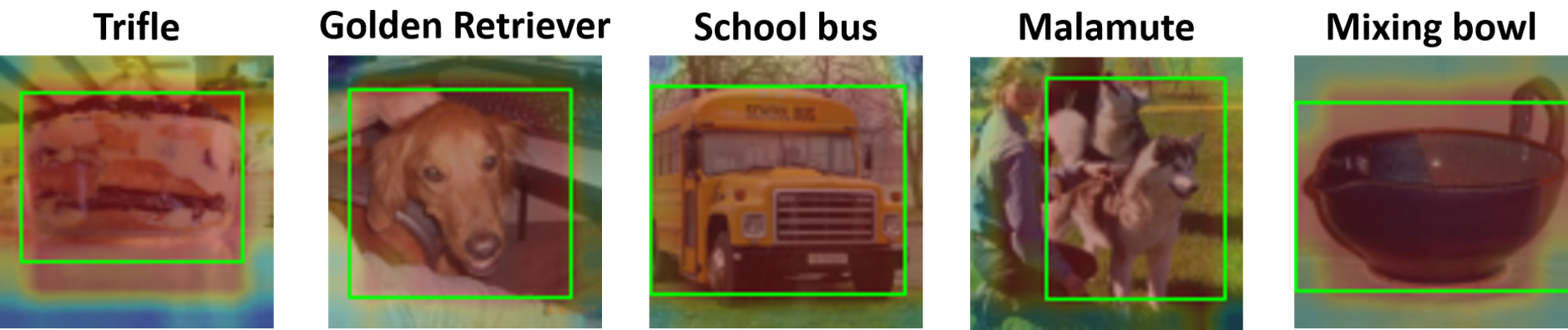}}
\vspace{-0.2cm}
\caption{Localization of foreground objects in support images in a \textit{mini}ImageNet 5-way 1-shot scenario.}
\label{fig:4}
\end{figure}

We further examined if our framework can localize the class object even when a single support image is given. The class object can be inferred as the common object \textit{among} the support images that share the same label. In 1-shot scenarios, however, we cannot know which object the label indicates without additional information.
Nevertheless, our model finds the region most highly correlated to its overall feature and gives plausible predictions when the foreground objects are large and significant. Figure \ref{fig:4} shows some of the success cases of the 1-shot foreground object localization.

\section{Conclusion}
\label{sec:print}
In this paper, we investigate the representation learning in the few-shot realm and point out that average pooling of the whole image features can result in impure representations. To overcome this limitation, we propose a framework that localizes the class-relevant regions of the images with weak supervision using class prototypes and extracts a refined class representation for each class. We also introduce a loss function that directly addresses the refined features. Our approach outperforms the baseline work in various benchmarks. It also shows reasonable localization results without bounding box annotations.
\pagebreak
\bibliographystyle{IEEEbib}
\bibliography{refs}

\end{document}